\documentclass[acmsmall,screen,nonacm]{acmart}

\AtBeginDocument{%
  }

\setcopyright{rightsretained}
\copyrightyear{2026}
\acmYear{2026}

\title{Frontier Large Language Models Rival State-of-the-Art Planners}

\author{Augusto B. Corrêa}
\affiliation{%
  \institution{University of Oxford}
  \city{Oxford}
  \country{United Kingdom}
}
\email{augusto.blaascorrea@chch.ox.ac.uk}

\author{André G. Pereira}
\affiliation{%
  \institution{Federal University of Rio Grande do Sul}
  \city{Porto Alegre}
  \country{Brazil}
}
\email{agpereira@inf.ufrgs.br}

\author{Jendrik Seipp}
\affiliation{%
  \institution{Linköping University}
  \city{Linköping}
  \country{Sweden}
}
\email{jendrik.seipp@liu.se}

\usepackage{booktabs}
\usepackage{bm}
\usepackage{csquotes}
\usepackage{enumitem}
\usepackage[draft,newfloat]{minted}
\usepackage{nicefrac}
\usepackage{pgfplots}
\usepackage{pgf}
\usepackage{subcaption}
\usepackage{todonotes}
\usepackage{comment}
\usepackage{verbatimbox}

\usepackage{tikz}
\usepackage{graphicx}
\usepackage{multirow}
\pgfplotsset{compat=1.18}
\usetikzlibrary{patterns}
\usetikzlibrary{arrows.meta, positioning, calc}

\setminted{fontsize=\scriptsize}

\newcommand{\mypar}[1]{\paragraph*{#1.}}

\renewcommand{\cite}[1]{\citep[][]{#1}}
\newcommand{\egcite}[1]{\citep[e.g.,][]{#1}}

\definecolor{mycolor0}{rgb}{0.12156862745098039, 0.4666666666666667, 0.7058823529411765}
\definecolor{mycolor1}{rgb}{1.0, 0.4980392156862745, 0.054901960784313725}
\definecolor{mycolor2}{rgb}{0.17254901960784313, 0.6274509803921569, 0.17254901960784313}
\definecolor{mycolor3}{rgb}{0.8392156862745098, 0.15294117647058825, 0.1568627450980392}
\definecolor{mycolor4}{rgb}{0.5803921568627451, 0.403921568627451, 0.7411764705882353}
\definecolor{mycolor5}{rgb}{0.5490196078431373, 0.33725490196078434, 0.29411764705882354}
\definecolor{mycolor6}{rgb}{0.8901960784313725, 0.4666666666666667, 0.7607843137254902}
\definecolor{mycolor7}{rgb}{0.4980392156862745, 0.4980392156862745, 0.4980392156862745}
\definecolor{mycolor8}{rgb}{0.7372549019607844, 0.7411764705882353, 0.13333333333333333}
\definecolor{mycolor9}{rgb}{0.09019607843137255, 0.7450980392156863, 0.8117647058823529}

\begin{document}

\begin{abstract}
A series of influential studies established that large language models cannot
reliably solve even simple planning tasks. We show that the latest generation of
frontier models overturns this conclusion. We evaluate three families of
frontier LLMs on a challenging set of planning tasks based on the most recent
International Planning Competition following rigorous evaluation guidelines:
solutions are verified with a validation tool, tasks are freshly created to
avoid data contamination, and performance is compared against state-of-the-art
classical planners. On standard task descriptions, Gemini 3.1 Pro outperforms
the strongest planner baseline (245 vs.\ 234 solved tasks out of 360), while
GPT-5 achieves comparable performance to the baselines. When all semantic
information is obfuscated from the descriptions to test for pure symbolic
planning, performance degrades but Gemini 3.1 Pro remains competitive with the
strongest baselines. A longitudinal comparison across model generations --- from
GPT-3.5, which solves zero tasks, to GPT-5 --- reveals a striking upward
trajectory. Frontier LLMs might finally be able to plan; the question now is how
far this capability will extend.
\end{abstract}


\maketitle

\noindent\setlength{\fboxrule}{0.5pt}\setlength{\fboxsep}{10pt}%
\fbox{\parbox{0.93\columnwidth}{%
\textbf{\large Key Insights}
\begin{itemize}[leftmargin=1.2em, topsep=4pt, itemsep=4pt]
\item Frontier LLMs have made striking progress on automated planning. On standard task descriptions from the International Planning Competition (IPC), the best model, Gemini~3.1~Pro, outperforms state-of-the-art planners, while GPT-5 achieves competitive performance.
\item When all semantic information is stripped from task descriptions through symbol obfuscation, LLM performance still degrades. Yet, Gemini~3.1~Pro remains competitive with the IPC winners even in this adversarial setting.
\item A longitudinal comparison across model generations reveals a clear upward trajectory, with the 2025 generation of LLMs not only closing the gap with specialized planners but surpassing them on a challenging set of benchmarks.
\end{itemize}
}}
\medskip

In the last few years, several studies argued that large language models (LLMs)
cannot plan \cite{valmeekam-et-al-neurips2023,dziri-et-al-neurips2023,kambhampati-blogcacm2023,kambhampati-et-al-icml2024,bo-et-al-arxiv2024,kokel-et-al-aaai2025,huang-et-al-icaps2025}.
LLMs, the argument goes, are sophisticated pattern matchers: they generate
fluent text but cannot reason through the kind of multi-step problem solving
that planning requires. Techniques such as Chain-of-Thought \cite{wei-et-al-neurips2022} prompting and
fine-tuning did not change this picture
\cite{stechly-et-al-neurips2024,rossetti-et-al-icaps2024,huang-et-al-icaps2025},
and even reasoning-oriented models incurred far greater computational cost
than classical planners without matching their performance
\cite{correa-et-al-neurips2025}.

We show that the latest generation of frontier LLMs has changed this landscape.
On a challenging set of planning benchmarks from the Learning track of the most recent International
Planning Competition (IPC) \cite{taitler-et-al-aimag2024}, Gemini~3.1~Pro
outperforms the winners of the last IPC (234 tasks solved), solving 245 tasks in total. Gemini~3~Pro and GPT-5 \cite{openai-tr2025} also achieve comparable
coverage to well-established planners. Even when all semantic information is obfuscated
from the task descriptions---replacing meaningful names with random strings---Gemini~3.1~Pro remains competitive with state-of-the-art planners. A
longitudinal comparison across model generations, from GPT-3.5 (which solves
zero tasks in our benchmark) to GPT-5, reveals a striking upward trajectory.

Automated planning---finding a sequence of actions that transforms an initial
state into a goal state---is a particularly rigorous benchmark for reasoning.
Tasks are formally specified, so solutions can be verified automatically \egcite{howey-long-icaps2003wscompetition}.
Difficulty scales from trivial to extremely hard, and the diversity of
available domains---from puzzle-solving to elevator control and manufacturing
\cite{hoffmann-et-al-jair2006,koehler-schuster-aips2000,liporace-et-al-ipc2004,long-fox-jair2003}---means
that strong performance signals genuine reasoning ability rather than simply
task memorization. These properties make automated planning an ideal testbed, and one on which LLMs have, until now, consistently failed.

\pgfplotsset{compat=1.11,
/pgfplots/ybar legend/.style={
    /pgfplots/legend image code/.code={%
    \draw[##1,/tikz/.cd,yshift=-0.25em]
    (0cm,0cm) rectangle (10pt,0.8em);},
    },
    }

\begin{figure}[t]
    \centering
\scalebox{0.86}{
\begin{tikzpicture}
\begin{axis}[
    ybar,
    width=10cm,
    height=5cm,
    bar width=20pt,
    ymin=0, ymax=285,
    ylabel={Solved Tasks},
    symbolic x coords={Standard Planning Tasks, Obfuscated Planning Tasks},
    xtick=data,
    enlarge x limits=0.525,
    ymajorgrids,
    grid style=dashed,
    nodes near coords,
    nodes near coords align={vertical},
    tick label style={font=\small},
    ylabel style={font=\small},
    legend style={
        at={(0.5,-0.3)},
        anchor=north,
        legend columns=-1,
        font=\small,
        /tikz/every even column/.append style={column sep=6pt}
    },
    xtick style={draw=none},
    ytick style={draw=none},
]

\addplot[
    pattern=north east lines,
    pattern color=blue!70!black
] coordinates {
    (Standard Planning Tasks,234)
    (Obfuscated Planning Tasks,234)
};

\addplot[
    fill=blue!35
] coordinates {
    (Standard Planning Tasks,157)
    (Obfuscated Planning Tasks,93)
};

\addplot[
    fill=orange!60
] coordinates {
    (Standard Planning Tasks,205)
    (Obfuscated Planning Tasks,152)
};

\addplot[
    fill=mycolor3
] coordinates {
    (Standard Planning Tasks,245)
    (Obfuscated Planning Tasks,231)
};

\legend{
     Planner,
     DeepSeek-R1,
     GPT-5,
     Gemini 3.1 Pro
}

\end{axis}
\end{tikzpicture}
}
\caption{End-to-end planning performance of frontier LLMs and a classical planner (Scorpion Maidu, the IPC 2023 Satisficing track winner) on standard and obfuscated planning tasks from eight domains of the IPC 2023 Learning Track.}
\label{fig:planning_tasks}
\end{figure}

In this article, we evaluate the end-to-end planning capabilities of three families of frontier LLMs (DeepSeek-R1 \cite{deepseek-arxiv2025}, OpenAI's GPT series \cite{openai-tr2025}, and Google's Gemini \cite{gemini-arxiv2025,google-gemini3}) following recently proposed guidelines for LLM planning research \cite{katz-et-al-arxiv2025}: all plans are verified with the sound validation tool VAL \cite{howey-long-icaps2003wscompetition}, tasks are newly generated to reduce the risk of data contamination, and we compare against strong planning baselines, including popular planners \cite{richter-westphal-jair2010} and the winners of the most recent IPC \cite{correa-et-al-ipc2023c,gnad-et-al-ipc2023b}.
We focus on \emph{satisficing planning}, where any valid plan is acceptable, as
the most natural setting for LLMs, which lack optimization guarantees.
Extending to optimal, temporal, or numeric planning remains
an open challenge.
Figure~\ref{fig:planning_tasks} summarizes our main findings. The frontier models can produce remarkably long valid plans of several hundred steps. These results mark a notable shift: the strongest frontier LLMs have not only closed the gap with specialized planners, but surpassed them on challenging benchmarks.
This comes at a steep price: while the planning baselines can solve some of the tasks in seconds on a single CPU, LLM inference requires massive GPU clusters and orders of magnitude more energy. Whether this trade-off is worthwhile depends on the application, but the emerging planning capability of frontier models is new.

\section{Background: Planning}

Planning tasks are typically described using the Planning Domain Definition Language (PDDL) \cite{mcdermott-aimag2000,haslum-et-al-2019}, a standard formalism in the AI planning community.\footnote{For a detailed treatment of PDDL semantics, see \citet[][Section~2]{helmert-aij2009}.} A PDDL specification consists of two parts: a \emph{domain file}, which defines the types of objects, their possible relationships (called predicates), and the available actions; and a \emph{task file}, which specifies the concrete objects, the initial state of the world, and the goal to be achieved.

Consider a simple logistics scenario as an example. The domain defines that trucks can be at different cities, packages can be loaded onto trucks, and a ``drive'' action moves a truck between connected cities. A particular task then specifies the concrete objects and initial state---e.g., a truck in Paris carrying a package---and the goal, e.g. deliver the package to London. A \emph{plan}---load the package, drive to London, unload---is a sequence of actions that transforms the initial state into one satisfying the goal. A \emph{planner} is the algorithm that finds such a plan. Figure~\ref{fig:pddl-example} shows the PDDL encoding of the ``drive'' action.

\begin{verbbox}
(:action drive
 :parameters (?t - truck ?from ?to - city)
 :precondition (and (at ?t ?from)
                    (connected ?from ?to))
 :effect (and (at ?t ?to)
              (not (at ?t ?from))))
\end{verbbox}
\begin{figure}[t]
    \centering
    \theverbbox
\caption{PDDL encoding of a \texttt{drive} action in a logistics domain. The action takes a truck from one city to a connected city, updating its location. This is the kind of formal specification that each LLM receives as input in our experiments.}
\label{fig:pddl-example}
\end{figure}

We focus on \emph{satisficing planning}, where any valid plan is acceptable and optimality is not required. The primary evaluation metric is \emph{coverage}---the number of tasks for which a valid plan is found---though we also examine \emph{plan length} as an indicator of solution quality. Unlike classical planners, LLM-based planning is neither guaranteed to produce only valid plans (\emph{soundness}) nor to find a plan whenever one exists (\emph{completeness}). We address the soundness concern by validating all LLM-generated plans with VAL \cite{howey-long-icaps2003wscompetition}, discarding any plans that fail verification.

\section{Evaluating Frontier LLMs}

We evaluate models from three families: DeepSeek-R1 \cite{deepseek-arxiv2025}, a reasoning-oriented model; OpenAI's GPT-3.5 \cite{openai-gpt35}, GPT-4.1 \cite{openai-gpt41}, and GPT-5 \cite{openai-tr2025}; and Google's Gemini~2.5~Pro \cite{gemini-arxiv2025}, Gemini~3~Pro \cite{google-gemini3}, and Gemini~3.1~Pro \cite{google-gemini31}. We refer to DeepSeek-R1, GPT-5, and Gemini~3.1~Pro as \emph{frontier models}, as they are the strongest available model in each family at the time of our experiments. All experiments use official APIs with default parameters and no tool use enabled; Appendix~\ref{app:experimental-details} lists the exact model identifiers and decoding settings. The experimental data (source code, benchmarks, logs, and evaluations) will be made publicly available online upon acceptance.

\mypar{Benchmark design} Our benchmark comprises eight domains from the Learning Track of the 2023 International Planning Competition (IPC) \cite{taitler-et-al-aimag2024}, covering a range of reasoning challenges from puzzle-solving (Blocksworld, Sokoban) to logistics and resource management (Rovers, Transport). This provides a representative selection of classical planning difficulty. We generate 45 novel tasks per domain using the IPC test set's parameter distributions to mitigate data contamination. These tasks are substantially harder than those typically used to evaluate LLMs on planning: for example, the commonly used PlanBench dataset \cite{valmeekam2023planbench} contains Blocksworld tasks with at most 5 blocks and optimal plans of at most 16 steps, whereas our benchmark set includes Blocksworld tasks with up to 477 blocks and plans exceeding 1\,000 steps.
Appendix~\ref{app:domains} describes each domain and lists the full parameter distributions.

Each LLM receives a prompt containing the PDDL domain and task files, two illustrative examples from other domains (each with a sample task and valid plan), and a checklist of common pitfalls \cite{correa-et-al-neurips2025}.
Appendix~\ref{app:prompt} describes the prompt in detail. As our planning baselines, we use three state-of-the-art classical planners with a 30-minute time limit and 8\,GiB memory per task: LAMA \cite{richter-westphal-jair2010}, a common baseline in prior LLM planning studies \egcite{silver-et-al-aaai2024};

DecStar \cite{gnad-et-al-ipc2023b}, the winner of the IPC 2023 Agile track; and Scorpion Maidu \cite{correa-et-al-ipc2023c}, the winner of the IPC 2023 Satisficing track. We tried other planners during our experiments, but Scorpion Maidu and DecStar were the strongest ones.  To test whether LLMs reason over the formal structure or simply exploit familiar tokens, we also create obfuscated versions of every domain and task: all action names, predicates, and object names are replaced with random strings \cite{valmeekam-et-al-neurips2023,chen-et-al-icaps2025wslm4plan}. This obfuscation is highly adversarial for LLMs---which rely on token semantics---but does not affect symbolic planners.

\begin{table*}[t]
\centering
\begin{tabular}{l r r r r r r r r r r r r}
                 & \multicolumn{3}{c}{Planners} & & \multicolumn{7}{c}{LLMs}  \\
 \cmidrule(lr){2-4} \cmidrule(lr){6-12}
                 & \multirow{2}{*}{\rotatebox{90}{LAMA}} & \multirow{2}{*}{\rotatebox{90}{DecStar}} & \multirow{2}{*}{\rotatebox{90}{Maidu}} & & DeepSeek & \multicolumn{3}{c}{OpenAI GPT} &  \multicolumn{3}{c}{Gemini Pro} \\
 \cmidrule(lr){6-6} \cmidrule(lr){7-9} \cmidrule(lr){10-12}
                 &             &             &             &  & R1  & 3.5 & 4.1 & 5.0         & 2.5 & 3.0         & 3.1          \\[3pt] \midrule
Blocksworld (45) & 28          & 26          & 29          &  & 19  & 0   & 3   & 21          & 28  & \textbf{35} & 34           \\
Childsnack (45)  & 18          & 31          & 23          &  & 33  & 0   & 6   & 38          & 41  & 41          & \textbf{44}  \\
Floortile (45)   & 10          & 10          & 21          &  & 5   & 0   & 0   & 10          & 4   & \textbf{26} & 25           \\
Miconic (45)     & \textbf{45} & \textbf{45} & \textbf{45} &  & 27  & 0   & 6   & 35          & 20  & 31          & 36           \\
Rovers (45)      & \textbf{34} & 28          & 31          &  & 9   & 0   & 0   & 14          & 6   & 11          & 16           \\
Sokoban (45)     & 21          & 17          & \textbf{22} &  & 10  & 0   & 0   & 15          & 11   & 14          & 15           \\
Spanner (45)     & 15          & 38          & 30          &  & 38  & 0   & 3   & \textbf{45} & 31  & \textbf{45} & \textbf{45}  \\
Transport (45)   & \textbf{33} & 32          & \textbf{33} &  & 16  & 0   & 2   & 27          & 20  & 30          & 30           \\ \midrule
Sum (360)        & 204         & 227         & 234         &  & 157 & 0   & 20  & 205         & 161 & 233         & \textbf{245} \\
\bottomrule                                                                                                                      \\
\end{tabular}
\caption{End-to-end planning performance of frontier large language models and
  classical planners on \textbf{standard planning domains} from the IPC
  2023 Learning Track. LAMA is the standard baseline used in prior LLM planning
  studies; DecStar and Scorpion Maidu are the winners of the IPC 2023 Agile and
  Satisficing tracks, respectively. The best performance for each domain is highlighted in
  bold.}
\label{table:coverage-standard}
\end{table*}

\begin{table*}[t]
\centering
\begin{tabular}{l r r r r r r r r r r r r}
                 & \multicolumn{3}{c}{Planners} & & \multicolumn{7}{c}{LLMs}  \\
 \cmidrule(lr){2-4} \cmidrule(lr){6-12}
                 & \multirow{2}{*}{\rotatebox{90}{LAMA}} & \multirow{2}{*}{\rotatebox{90}{DecStar}} & \multirow{2}{*}{\rotatebox{90}{Maidu}} &   & DeepSeek & \multicolumn{3}{c}{OpenAI GPT} &  \multicolumn{3}{c}{Gemini Pro} \\
 \cmidrule(lr){6-6} \cmidrule(lr){7-9} \cmidrule(lr){10-12}
                 &             &             &              &  & R1 & 3.5 & 4.1 & 5.0 & 2.5 & 3.0         & 3.1         \\[3pt] \midrule
Blocksworld (45) & 28          & 26          & 29           &  & 3  & 0   & 0   & 12  & 25  & \textbf{39} & 31          \\
Childsnack (45)  & 18          & 31          & 23           &  & 28 & 0   & 0   & 33  & 29  & 31          & \textbf{40} \\
Floortile (45)   & 10          & 10          & \textbf{21}  &  & 0  & 0   & 0   & 3   & 2   & 13          & 18          \\
Miconic (45)     & \textbf{45} & \textbf{45} & \textbf{45}  &  & 14 & 0   & 0   & 28  & 10   & 23          & 35          \\
Rovers (45)      & \textbf{34} & 28          & 31           &  & 5  & 0   & 0   & 10  & 6   & 13          & 15          \\
Sokoban (45)     & 21          & 17          & \textbf{22}  &  & 0  & 0   & 0   & 7   & 6   & 11          & 14          \\
Spanner (45)     & 15          & 38          & 30           &  & 34 & 0   & 0   & 38  & 41  & \textbf{45} & 44          \\
Transport (45)   & 33          & 32          & 33           &  & 9  & 0   & 0   & 21  & 23  & 28          & \textbf{34} \\ \midrule
Sum (360)        & 204         & 227         & \textbf{234} &  & 93 & 0   & 0   & 152 & 142 & 203         & 231         \\
\bottomrule                                                                                                             \\
\end{tabular}
\caption{End-to-end planning performance of frontier large language models and
  classical planners on \textbf{obfuscated planning domains} from the IPC
  2023 Learning Track. Obfuscation does not affect classical planners, so their
  results are identical to Table~\ref{table:coverage-standard}. The best performance for each domain is highlighted in
  bold.}
\label{table:coverage-obfuscated}
\end{table*}

\mypar{Performance on standard tasks} Tables~\ref{table:coverage-standard} and \ref{table:coverage-obfuscated} present the full results. On standard tasks, Gemini~3.1~Pro leads with 245 solved tasks out of 360, surpassing not only LAMA (204) but also the IPC 2023 winners DecStar (227) and Scorpion Maidu (234). Gemini~3~Pro solves 233 tasks, GPT-5 closely matches LAMA at 205, while DeepSeek-R1 and Gemini~2.5~Pro solve 157 and 161 tasks, respectively. LLMs particularly excel in domains with repetitive action patterns: in both Childsnack and Spanner, all three frontier models match or outperform every classical planner. In Spanner, GPT-5, Gemini~3~Pro, and Gemini~3.1~Pro all solve all 45 tasks. Gemini~3.1~Pro performs consistently well across domains, outperforming Scorpion Maidu in Blocksworld (34 vs.\ 29), Childsnack (44 vs.\ 23), Floortile (25 vs.\ 21), and Spanner (45 vs.\ 30).

\mypar{Role of semantic information} The obfuscated tasks reveal the extent to which LLMs rely on the meaning of symbols. Gemini~3.1~Pro drops from 245 to 231 solved tasks ($-5.7\%$), remaining close to Scorpion Maidu (234). Gemini~3~Pro drops from 233 to 203 ($-12.9\%$). GPT-5 shows a larger decline from 205 to 152 tasks ($-25.9\%$), and DeepSeek-R1 drops from 157 to 93 ($-40.8\%$). GPT-4.1, which solved 20 standard tasks, fails entirely on obfuscated ones. The fact that Gemini~3.1~Pro still solves 231 of these purely symbolic tasks---nearly matching Scorpion Maidu---is a striking result, given that some of the early LLMs could not solve a single task in the obfuscated setting.

\mypar{Information leakage} Since Gemini models expose reasoning tokens, we can inspect the model's chain of thought for signs of data contamination.
In the obfuscated setting, Gemini~3.1~Pro's traces reveal that it frequently identifies the underlying PDDL domain from structural cues alone---for example, recognizing a grid-based push-box structure as Sokoban even when every symbol has been replaced.
This re-identification occurs consistently for Blocksworld, Floortile, Sokoban and Transport, but never for Childsnack, Miconic, Rovers or Spanner.
For the four recognized domains, some portion of the model's performance may therefore be aided by prior exposure to similar tasks during training, rather than reflecting pure symbolic reasoning.
However, this does not fully explain Gemini~3.1~Pro's strong obfuscated performance: on the four domains the model does \emph{not} identify---Childsnack, Miconic, Rovers, and Spanner---it still solves the vast majority of tasks it solved in the standard setting, losing only 4, 1, 1 and 1 tasks, respectively.
This suggests that the model's capability is not reducible to domain recall.%

\mypar{A longitudinal perspective} The progression within the GPT family is particularly instructive. GPT-3.5 cannot solve a single task in our benchmark. GPT-4.1 manages 20 standard tasks but fails on all obfuscated ones. GPT-5 then solves 205 standard and 152 obfuscated tasks---a dramatic leap that can likely be attributed to its specific training for problems that require reasoning~\cite{openai-tr2025}. This generational trajectory, from zero performance to competitiveness with state-of-the-art planners, illustrates the rapid pace of improvement.

\begin{figure*}[t!]
\centering
\begin{subfigure}[t]{0.48\textwidth}
    \centering
    \scalebox{0.36}{\input{plan-lengths-jitter.pgf}}
    \caption{Distribution of plan lengths for tasks solved by each LLM. GPT-5
      found one plan with 1\,729 steps (Miconic domain), which is
      omitted from the plot.}
    \label{fig:plan-lengths}
\end{subfigure}
\hfill
\begin{subfigure}[t]{0.48\textwidth}
    \centering
    \scalebox{0.3}{\input{plot-tokens.pgf}}
    \caption{Number of reasoning tokens used by Gemini 3.1 Pro to solve tasks in the Standard and Obfuscated settings.}
    \label{fig:plot-tokens}
\end{subfigure}
\caption{Plan lengths and reasoning effort. (a) Frontier LLMs can generate valid plans exceeding 200 steps. (b) Gemini~3.1~Pro expends substantially more reasoning tokens on obfuscated tasks, yet the performance gap remains modest.}
\label{fig:plans-and-tokens}
\end{figure*}

\mypar{Plan length and reasoning effort} Figure~\ref{fig:plan-lengths} shows the distribution of plan lengths for solved tasks. Frontier LLMs can generate remarkably long valid plans, including several exceeding 200 steps. Maintaining such extended chains of correct reasoning represents a qualitative improvement over earlier generations, which struggled even with short plans \cite{valmeekam-et-al-neurips2023,valmeekam-et-al-arxiv2024}. Figure~\ref{fig:plot-tokens} offers a complementary perspective: for Gemini~3.1~Pro, the number of reasoning tokens increases substantially on obfuscated tasks compared to standard ones, sometimes even reaching the reasoning token limit of 65\,536 tokens. The model uses much more computational effort to compensate for the absence of semantic cues---yet the performance gap remains.
Since the Gemini models provide access to reasoning tokens, we can also analyze the strategies they use to solve tasks.
This reveals a mix of domain-specific approaches and general problem-solving techniques. For example, in Blocksworld, the model often adopts the well-known strategy of unstacking all blocks to the table and then building the goal configuration from scratch, rather than exploiting partial structures or interleaving unstacking with goal-directed stacking.

\section{Conclusions and Discussion}

Our results indicate that frontier LLMs have made substantial progress in automated planning. On standard PDDL tasks, Gemini~3.1~Pro outperforms state-of-the-art planners, including the winners of IPC 2023, and GPT-5 solves roughly as many tasks as LAMA. This represents a major shift from just two years earlier, when LLMs could not solve even the simplest planning tasks reliably \cite{valmeekam-et-al-neurips2023}.

The obfuscation experiments are particularly revealing. The consistent performance drop confirms that current LLMs still leverage semantic information when solving planning tasks---the meaning of action and predicate names provides useful context that helps the model solve the task. However, the fact that frontier models, especially Gemini~3.1~Pro, retain substantial capability even without semantic cues suggests that they can solve tasks that require reasoning without relying on explicit pattern matching over semantic information.
This is a qualitatively different picture from earlier evaluations, where obfuscation caused near-complete failure.

\mypar{Efficiency} This progress must be weighed against the enormous computational cost of LLM-based planning. The baseline planners run on a single CPU core with 8\,GiB of memory and typically solves tasks within seconds. The LLMs, by contrast, require massive GPU clusters for inference. DeepSeek-R1, a 671-billion-parameter Mixture-of-Experts model, alone requires over 1\,000\,GiB of GPU memory. Even when LLMs match or exceed a classical planner's coverage, they consume orders of magnitude more energy and wall-clock time---a decisive consideration for any practical planning application.

\mypar{Outlook} The rapid improvement shown in our longitudinal comparison naturally raises the question of how far this progress will extend. Several challenges remain. Our benchmark targets satisficing planning, where any valid plan suffices; optimal planning, which requires a shortest plan, is a fundamentally harder problem, and classical planners still hold a decisive advantage there. To our knowledge, no fully generative method guarantees optimality, though many neuro-symbolic approaches exist. A promising direction is to combine the flexibility of LLMs---their ability to handle natural language, incomplete specifications, and novel domains without engineering effort---with the formal guarantees and scalability of classical planners. Perhaps most intriguingly, understanding \emph{why} reasoning-capable LLMs have acquired planning ability could shed light on the nature of planning itself. Our results suggest that the question ``can LLMs plan?'' is settling; the more productive question is now: what should we build with planning-capable LLMs?

\begin{acks}
This work was supported by the Wallenberg AI, Autonomous Systems and Software Program (WASP) funded by the Knut and Alice Wallenberg Foundation.
André G.~Pereira acknowledges support from FAPERGS with projects 21/2551-0000741-9 and 25/2551-0002590-7.
This study was financed in part by the \textit{Coordenação de Aperfeiçoamento de Pessoal de Nível Superior -- Brasil} (CAPES) -- Finance Code~001.
\end{acks}

\bibliography{abbrv,literatur,additional,crossref}

\appendix

\section{Benchmark Domains}
\label{app:domains}

Our benchmark uses eight domains from the Learning Track of the 2023 International Planning Competition \cite{taitler-et-al-aimag2024}. Table~\ref{tab:domain-params} summarizes the task parameters and the longest plans found by LAMA. Below we briefly describe each domain.

\begin{description}
  \item[Blocksworld.] Rearrange a set of $n$ blocks on a table into a goal configuration by stacking and unstacking them one at a time.
  \item[Childsnack.] Prepare and serve sandwiches to $c$ children at different tables in a kitchen-logistics setting while respecting the allergy constraints of each child.
  \item[Floortile.] A set of robots paint $t$ floor tiles in a specified pattern, moving between adjacent cells with certain restrictions and switching paint colors.
  \item[Miconic.] An elevator must transport $p$ passengers to their destination floors, picking up and dropping off passengers along the way.
  \item[Rovers.] Coordinate $r$ planetary rovers to navigate waypoints, collect soil and rock samples, take images, and transmit data.
  \item[Sokoban.] An agent pushes $b$ boxes onto designated goal locations in a grid. Boxes can only be pushed (not pulled).
  \item[Spanner.] An agent collects $s$ spanners distributed across locations and uses them to tighten nuts at a gate.
  \item[Transport.] Use $v$ vehicles to pick up and deliver packages across a network of locations with varying distances.
\end{description}

\begin{table}[t]
\centering
\begin{tabular}{lrr}
\textbf{Domain} & \textbf{Parameters} & \textbf{Max. Length} \\ \midrule
Blocksworld     & $n \in [5, 477]$    & 1194                 \\
Childsnack      & $c \in [4, 284]$    & 252                  \\
Floortile       & $t \in [12, 899]$   & 62                   \\
Miconic         & $p \in [1, 470]$    & 1438                 \\
Rovers          & $r \in [1, 29]$     & 1194                 \\
Sokoban         & $b \in [1, 78]$     & 860                  \\
Spanner         & $s \in [1, 474]$    & 21                   \\
Transport       & $v \in [3, 49]$     & 212                  \\
\bottomrule \\
\end{tabular}
\caption{Task size for each standard domain based on their main parameters: $n$ blocks in
  Blocksworld, $c$ children in Childsnack, $t$ tiles in Floortile, $p$
  passengers in Miconic, $r$ rovers in Rovers, $b$ boxes in Sokoban, $s$
  spanners in Spanner, and $v$ vehicles in Transport. We also show the longest
  (and possibly suboptimal) plan found by the baseline planner LAMA.}
\label{tab:domain-params}
\end{table}

\section{Experimental Details}
\label{app:experimental-details}

The model identifiers used are: \texttt{deepseek-reasoner}, \texttt{gpt-3.5-turbo-0125}, \texttt{gpt-4.1-2025-04-14}, \texttt{gpt-5-2025-08-07}, \texttt{gemini-2.5-pro}, \texttt{gemini-3-pro-preview}, and \texttt{gemini-3.1-pro-preview}.

We use temperature $0.1$ and top-$p$ of $0.5$ for all models, except for GPT-5 which only accepts a temperature of $1.0$. In particular, we do not adjust top-$p$ or other sampling settings. The maximum output token limit is not set, and we use the provider's default maximum output length in all cases.

Each model receives a single attempt per task. A task is considered solved if the model's output contains a plan that passes VAL verification. We retried runs only in case of network problems or rate limit errors.

\section{Prompt}
\label{app:prompt}

We use the same prompt as \citet{correa-et-al-neurips2025} to keep consistency with the literature. For different tasks, only three items change: the name of the domain, the \texttt{domain-file}, and \texttt{instance-file-1}. To reduce the length of our appendix, we do not display the entire domain and instance files, but just the first lines of each. We also show only the first few actions of each example plan. We use Blocksworld as an example below.

\begin{minted}{text}
<problem-description>
You are a highly-skilled professor in AI planning searching a plan for a PDDL task from the domain <domain>blocksworld</domain>. You will be given the PDDL domain and the PDDL instance, and you need to return the plan in the format shown below. Next, you will receive a sequence of examples for your task and finally the definition of the blocksworld domain.
</problem-description>

This is the PDDL domain file of the blocksworld domain:
<domain-file>
(define (domain blocksworld)
[...]
</domain-file>

This is the PDDL instance file, for which you need to find a plan:
<instance-file-1>
(define (problem blocksworld-01)
[...]
</instance-file-1>

This is the PDDL domain file of another domain, called Gripper, which serves as an example:
<gripper-domain-file>
[...]
</gripper-domain-file>

This is an example of an instance file from the Gripper domain:
<gripper-instance-file-example>
[...]
</gripper-instance-file-example>

This is a plan for the Gripper instance above:
<plan-gripper>
[...]
</plan-gripper>

This is the PDDL domain file of another domain, called Logistics, to serve as a second example:
<logistics-domain-file>
[...]
</logistics-domain-file>

This is an example of an instance file from the Logistics domain:
<logistics-instance-file-example>
[...]
</logistics-instance-file-example>

This is a plan for the Logistics instance above:
<plan-logistics>
[...]
</plan-logistics>

Provide only the plan for the given instance. Here is a checklist to help you with your task:
1) The plan must be in the same format as the examples above.
2) Use the tags "<plan>...</plan>" around the plan.
3) The actions in the plan must be from the set of actions in the domain blocksworld, that is, they must use the same name and same number of parameters as one of the action schemas.
4) The plan must be valid, that is, each action must be applicable in the state it is applied in and the plan must end in a goal state.
\end{minted}

\end{document}